\documentclass[11pt]{article}

\def\a{\mathbf{a}}

\usepackage{amsmath,amsthm}
\usepackage{microtype}
\usepackage{graphicx,subfigure}
\usepackage{algorithm,algpseudocode}

\newtheorem{theorem}{Theorem}[section]
\newtheorem{proposition}[theorem]{Proposition}

\begin{document}

\title{Improvements on ``Fast space-variant elliptical filtering using box splines''}
\author{Kunal~Narayan~Chaudhury and Sebanti~Sanyal
\thanks{Correspondence: K.~N.~Chaudhury (kchaudhu@math.princeton.edu). 
}}
\date{}

\maketitle

\begin{abstract}
It is well-known that box filters can be efficiently computed using pre-integrations and local finite-differences \cite{Crow1984,Heckbert1986,Viola2001}. 
By generalizing this idea and by combining it with a non-standard variant of the Central Limit Theorem, a constant-time or $O(1)$ algorithm 
was proposed in \cite{Chaudhury2010} that allowed one to perform space-variant filtering using Gaussian-like kernels . 
The algorithm was based on the observation that both isotropic and anisotropic Gaussians could be approximated using certain bivariate splines called \textit{box splines}.
The attractive feature of the algorithm was that it allowed one to continuously control the shape and size (covariance) of the filter, and that it had a fixed computational cost per pixel, irrespective of the size of the filter. 
The algorithm, however, offered a limited control on the covariance and accuracy of the Gaussian approximation. 
In this work, we propose some improvements by appropriately modifying the algorithm in \cite{Chaudhury2010}. 
\end{abstract}

\textbf{Keywords}:
Linear filtering, Gaussian approximation, Central limit theorem, Anisotropic Gaussian, Covariance, Box spline, Cartesian grid, Running sum, $O(1)$ algorithm.

\section{Introduction}

A space-variant filter is one where the shape and size of the filter is allowed to change from point-to-point within the image. Unlike the more standard convolution filter, the efficient computation of space-variant
filters (diffusion filters) remains a challenging problem in computer vision and image processing \cite{Weickert1994}. It has been shown that efficient algorithms for space-variant filtering can be designed 
using spline kernels, particularly when the space-variance is in terms of the scale (or size) of the kernel. For instance, Heckbert proposed an algorithm for adaptive image blurring using tensor-products of polynomial splines, 
where the image is filtered with kernels of various scales using repeated integration and finite-difference \cite{Heckbert1986}. Based on similar principles, namely, the scaling properties of B-splines, 
Munoz-Barrutia et al. have developed an algorithm for fast computation of the continuous wavelet transform at different scales \cite{Arrate2002}. The idea was extended in \cite{Arrate2010} to perform space-variant filtering using Gaussian-like functions 
of arbitrary size . This was done by approximating the Gaussian using separable B-splines. The flip side of using a separable construction, however, was that it offered limited steerability and ellipticity.
More recently, it was shown in \cite{Chaudhury2010} that this limitation could be fixed using certain non-separable splines 
called the \textit{box splines} \cite{deBoor}. This, however, did not solve the problem completely as the associated filtering algorithm offered only a limited control on the accuracy of the Gaussian approximation. 
In this paper, we address these algorithmic limitations and provide some simple solutions. To introduce the problem and to fix the notations, we briefly recall the main results in \cite{Chaudhury2010}. 
While we keep the presentation as self-contained as possible, we have avoided reproducing all the technical details, and refer the readers to original paper for a more complete account. 

\subsection{Background}

The key idea behind the fast $O(1)$ algorithm for Gaussian filtering in \cite{Chaudhury2010} is the use of a single global pre-integration, followed by local finite differences.
At the heart of this is the following non-standard form of the Central Limit Theorem\footnote{We refer the interested reader to \cite{ChaudhuryArxiv} for an interpretation of this result in the probabilistic setting.}. 

\begin{theorem}[Gaussian approximation, \cite{Chaudhury2010}] For a given integer $N \geq 2$, let $\theta_k=(k-1)\pi/N$ for $1 \leq  k \leq N$. Then 
\begin{equation}
\label{conv}
\lim_{N \longrightarrow \infty} \ \prod_{k=1}^N  \ \mathrm{sinc} \Bigg( \sigma \sqrt{\frac{6}{N}} ( x \cos \theta_k  + y \sin \theta_k) \Bigg) =\exp\Big(-\frac{\sigma^2}{2}(x^2+y^2)\Big).
\end{equation}
\end{theorem}
This result tells us that the Gaussian can be approximated by the product of rescaled sinc\footnote{We define $\mathrm{sinc}(t)$ to be $1$ at the origin and $\sin(t)/t$ otherwise.} functions that have been uniformly spread out over the half-circle. Note that each term in the product in \eqref{conv} is obtained 
through a rotation and rescaling of $\mathrm{sinc}(u) 1(v)$, the tensor-product between the $\mathrm{sinc}$ function and the constant function of unit height. The main idea behind \eqref{conv} is that $\mathrm{sinc}(t)$
behaves as $1-t^2/6+O(t^4)$ close to the origin, so that $[\mathrm{sinc}(t/\sqrt{N})]^N \approx \exp(-t^2/6)$ when $N$ is large compared to $t$.

\begin{figure}
\centering
\includegraphics[width=1.0\linewidth]{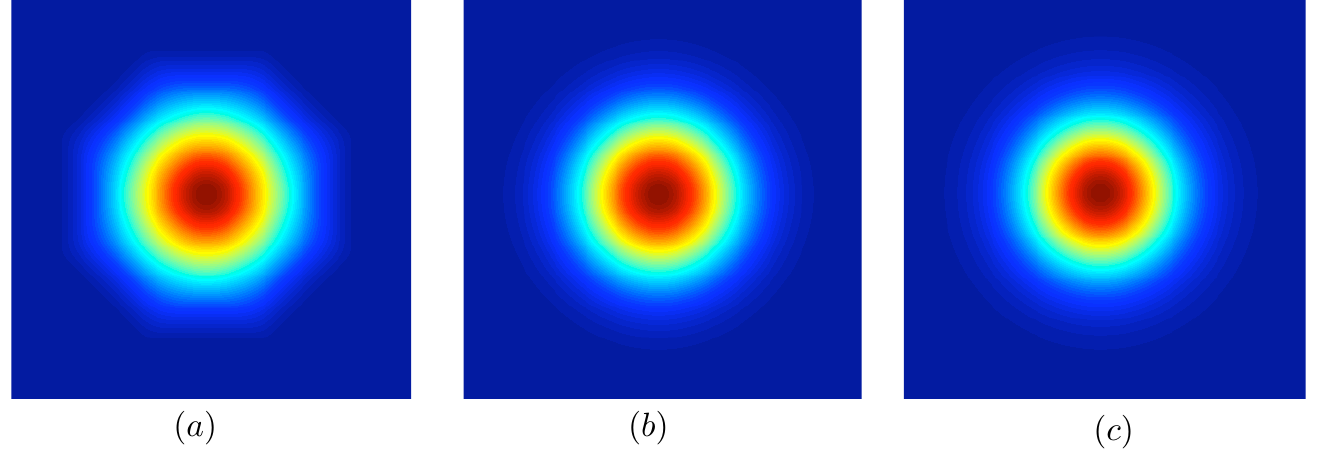}
\caption{Demonstration of the Central Limit Theorem used to approximate the isotropic Gaussian. (a) Convolution of four box distributions of equal width and uniformly distributed over $[0,\pi)$, (b) Convolution 
of eight box distributions of equal width and uniformly distributed over $[0,\pi)$, (c) Target Gaussian. The widths of the boxes are so adjusted that (a) and (b) have the same covariance as (c). Note that 
the convergence happens quite rapidly with the increase in the number of boxes. The maximum pointwise error between (b) and (c) is already within $1\%$ of the peak value.}
\label{figure4}
\end{figure}

So how exactly does this observation lead to an $O(1)$ algorithm for approximating Gaussian filters? By $O(1)$, we mean that the cost of filtering is independent of the size of the filter. It is clear that the right 
side of \eqref{conv} is simply the Fourier transform of the Gaussian 
\begin{equation*}
g(x,y)=\frac{1}{2\pi \sigma}\exp\Big(-\frac{x^2+y^2}{2\sigma^2}\Big).
\end{equation*}
On the other hand, the product in \eqref{conv} can be related to the box distribution $\mathrm{Box}_a(x) \delta(y)$, where 
\begin{equation*}
\mathrm{Box}_a(t)=
\begin{cases}  1/a &  \text{for \ $-a/2 <  t \leq  a/2$,} \\
0 &\text{otherwise},
\end{cases}
\end{equation*}
and  $\delta(y)$ is the delta function. Indeed, using the rotation-invariance and the multiplication-convolution property of the Fourier transform, we see that the left side of \eqref{conv} is
the Fourier transform of
\begin{align}
\label{boxSpline}
&\mathrm{Box}_{a_1}(x \cos \theta_1  + y \sin \theta_1) \delta(x \sin \theta_1 - y \cos \theta_1)\  \ast \cdots \nonumber \\
&\cdots \ast  \ \mathrm{Box}_{a_N}(x \cos \theta_N  + y \sin \theta_N) \delta(x \sin \theta_N - y \cos \theta_N),
\end{align}
where every $a_k$ equals $\sigma \sqrt{24/N}$. This gives us the dual interpretation of \eqref{conv} in the space domain, namely that $g(x,y)$ can be approximated using the convolution of
uniformly rotated box distributions. This idea is illustrated in Figure \ref{figure4}. The function in \eqref{boxSpline} turns out to be a compactly-supported piecewise
polynomial of degree $ \leq N - 2$, popularly called a \textit{box spline}. Note that we could as well replace the sinc function in \eqref{conv} by any
other ``bump'' function that looks like an inverted parabola around the origin, e.g. the cosine function as used in \cite{Chaudhury2011}. However, not every such function would be the Fourier transform of a simple, 
compactly-supported function such as the box function.

Applied to a continuously-defined function $f(x,y)$, this  means that we can approximate the Gaussian filtering $(f \ast g)(x,y)$ by successive convolutions with a fixed number of rotated box distributions. Now, note that 
the convolution of $f(x,y)$
with $\mathrm{Box}_a(x \cos \theta_k  + y \sin \theta_k) \delta(x \cos \theta_k-y \sin \theta_k)$ amounts 
to doing one-dimensional convolutions of $\mathrm{Box}_a(t)$ with profiles of $f(x,y)$ sampled along $L_{\theta}=\{(x,y): x \cos \theta_k  - y \sin \theta_k =  0\}$ and lines parallel to $L_{\theta}$. For example,
the convolution of $f(x,y)$ with $\mathrm{Box}_a(x) \delta(y)$ is given by one-dimensional convolutions of $\mathrm{Box}_a(t)$ with the profiles of $f(x,y)$ along the $x$-coordinate and lines parallel to it. 
The idea in \cite{Chaudhury2010} was to decompose the convolution with a box distribution into two steps:

\begin{itemize}
\item Pre-integration of the scan profiles along $L_{\theta_k}$ ($1 \leq k \leq N$) and lines parallet to it. This resulted in the so-called \textit{running sum}.
\item Application a finite-difference mesh to the running sum at each pixel, where the mesh parameters were determined by the scales $a_1,\ldots,a_N$.
\end{itemize}

By doing so, the cost of the convolutions with the box distributions was reduced to the cost of $(1)$ and $(2)$, which is clearly $O(1)$ per pixel irrespective of the values of $a_1,\ldots,a_N$. 

This idea was then extended to perform the more challenging space-variant filtering. This was based on the observation that by adjusting the scale of the box spline along each 
direction $\theta_1,\ldots,\theta_k$, one could control the anisotropy of the box spline. From the algorithmic perspective, this meant that we could change the filter at every pixel simply by changing the 
finite-difference mesh. This resulted in a fast algorithm for space-variant (or, non-convolution) filtering that had the same $O(1)$ complexity. 

In particular, for the case $N=4$, corresponding to $\theta_1 = 0, \theta_2 = \pi/4, \theta_3 = \pi/2$, $\theta_4 = 3\pi/4$ in \eqref{boxSpline}, a simple algorithm was developed that allowed one
to control the covariance simply by adjusting the scales $a_1,\ldots,a_4$ in \eqref{boxSpline}. We continue to use $\beta_{\a}(x,y)$ to denote this \textit{four-directional box spline}, where we call $\a = (a_1,a_2,a_3,a_4)$
the \textit{scale-vector}. The algorithm for the four-directional box spline proceeded in two steps. In the first step, the running sum was obtained by pre-integrating teh image along the 
directions $0, \pi/4, \pi/2$, and $3\pi/4$. This was done using fast recursions. To keep the running sum on the Cartesian grid, a step size of unity was used along the horizontal and vertical directions, 
while a step of $\sqrt{2}$ was used along the diagonals.
In the second step, a $16$-tap finite-difference mest was applied at point of the running sum, 
where the taps of the mesh was determined by the scale-vector. This produced the final filtered image. We refer the readers to Algorithm 1 in \cite{Chaudhury2010} for further details. Henceforth, we will refer to this as the \texttt{BoxFilter} algorithm.
More recently, similar ideas have also been applied in the field of computer
graphics \cite{Wetzstein2011}. By adapting the same algorithm, an efficient technique for detecting blobs in cell images was also  proposed in \cite{ChaudhuryISBI2010}.

The anisotropy of $\beta_{\a}(x,y)$ was specified using three parameters, its size ($s_{\a}$), elongation ($\rho_{\a}$), and orientation ($\theta_{\a}$). 
These were defined using the eigen decomposition of the covariance matrix of $\beta_{\a}(x,y)$,  
\begin{equation}
\label{cov}
\mathbf{C}_{\a} = \left(\begin{array}{cc}  \int \int x^2 \beta_{\a}(x,y) \ dx dy  &  \int \int xy  \beta_{\a}(x,y) \ dx dy \\
\int \int  xy \beta_{\a}(x,y) \ dx dy &   \int \int y^2 \beta_{\a}(x,y) \ dx dy\end{array}\right).
\end{equation}
The size was defined as the sum of the eigenvalues, the elongation as the ratio of the larger to the smaller eigenvalue, and the orientation as the angle (between $0$ and $\pi$) of the top eigenvector. 
In fact, the covariance of  $\beta_{\a}(x,y)$ was explicitly computed to be
\begin{equation}
 \mathbf{C}_{\a} = \frac{1}{24} \left(\begin{array}{cc} 2a^2_1+a^2_2+a^2_4 & a^2_2-a^2_4 \\
a^2_2-a^2_4 &  2a^2_3+a^2_2+a^2_4\end{array}\right).
\end{equation}
This resulted in the following formulas in terms of the scale-vector:
\begin{equation}
\label{geometry}
s_{\a}=\frac{1}{12}\sum a_k^2, \quad  \ \rho_{\a}= \frac{\sum a^2_k+\sqrt{D}}{\sum a^2_k-\sqrt{D}},  \text{ and }  \  \tan \theta_{\a}= \left(\frac{a^2_3-a^2_1+\sqrt{D}}{a^2_2-a^2_4}\right),
\end{equation}
where $D=(a^2_1-a^2_3)^2+(a^2_2-a^2_4)^2$. 

The scale vector was to control the anisotropy of the filter at every pixel, and this was done, in effect,
by controlling the tap positions of the finite-difference. Note that $\mathbf{C}_{\a}= (a^2/6) \mathbf{I}$ when all the four scales are equal to $a$. As a result, the best approximation of the isotropic Gaussian 
with variance $\sigma^2$ was obtained by seeting each scale to $\sqrt{6} \sigma$.  As for the anisotropic setting, a simple algorithm for uniquely determining the scales for a given size, elongation, and orientation (equivalently, 
for a given covariance) was provided in \cite{Chaudhury2010}. The algorithm took the covariance $\mathbf{C}$ as input and returned the box spline that had the minimum kurtosis among all box splines with covariance $\mathbf{C}$.

\subsection{Problem description}

There are, however, two bottlenecks with the above approach. The first is that one cannot control the accuracy of the Gaussian approximation in case of the space-variant filtering. 
We note that this problem  can be addressed for the more simple setting of convolution filtering simply using higher-order box splines (see Figure \ref{figure4}).
In effect, this amounts to repeated filtering of the image with the four-directional box spline
However, this cannot be done in the space-variant setting, where the kernel changes from point-to-point. 
The only possibility is to adapt the algorithm for the higher-order box splines. 
This, however, turns out to be of limited practical use since the size of the finite-difference mesh grows as $2^N$ with the order $N$ (two points per box distribution).

The other problem is the control on the elongation of the anisotropic box splines. This problem  is intimately related to the geometry of the Cartesian grid $\mathbf{Z}^2$. Note that the axes $L_{\theta}$ of the box distribution in the 
continuous setting corresponds to the discrete points $L_{\theta}=\{(m,n) \in \mathbf{Z}^2: m \cos \theta_k - n \sin \theta_k =  0\}$ on the grid. As a result, one requires $\tan \theta_k$ to be a rational number
to avoid interpolating the off-grid samples.  
It is, however, easily seen that, for $N>4$, one cannot find lines $L_{\theta_1},\ldots,L_{\theta_N}$ with the requirement that $\theta_1,\ldots,\theta_N$ are uniformly distributed. This restricted us 
to the four-directional box spline $\beta_{\a}(x,y)$. This restriction on the number of directions resulted in a ceiling on the maximum achievable elongation. In particular, while $\beta_{\a}(x,y)$ could arbitrarily elongated in the neighborhood of its four axes, 
it was difficult to achieve large elongations in the neighborhood of the mid-axes, i.e., along $\pi/8, 3\pi/8, 5\pi/8,$ and $7\pi/8$. It was found that, for every orientation $0 \leq \phi < \pi$, 
there is a bound $e(\phi)$ on the elongation achievable by $\beta_{\a}(x,y)$ given by
\begin{equation}
\label{boundE}
e(\phi) = \frac{1 + t + \sqrt{1+t^2}}{1 + t - \sqrt{1+t^2}}, 
\end{equation}
where $t = |\tan\phi - \cot \phi|/2$. This meant that $\rho_{\a}$ could be made at most as large as $e(\phi)$ under the constraint that $\theta_{\a} = \phi$, where $\rho_{\a}$ and $\theta_{\a}$ are as defined in \eqref{geometry}. 
The smallest value of $e(\phi)$ as $\phi$ varied over the half-circle was found to be roughly $5.8$, and this was reached along the mid-axes where one had the least control.

\section{Improved accuracy and elongation}

We now address the above mentioned problems for the four-directional box-spline. Our interest is mainly in the space-variant setting, where the filter varies from point-to-point in the image.

\subsection{Improved accuracy}
\label{problemA}

As remarked earlier, the Gaussian approximation can be improved by convolving the box spline with itself. All that needs to be guaranteed is that the covariance remains equal to that of the 
target Gaussian after the convolution. In fact, when the functions involved have some structure, one can exactly predict how the covariance changes with smoothing.

\begin{proposition}[Covariance of convolutions]
\label{covariance}
Let $f(x,y)$ and $g(x,y)$ be two functions which are symmetric around the origin, and which have unit mass. Then the covariance of $(f \ast g)(x,y)$ is the sum of the covariances of $f(x,y)$ and $g(x,y)$.
\end{proposition}

Note that $\beta_{\a}(x,y)$ automatically satisfies the conditions in the proposition. Our next observation is that we can get a higher-order space-variant filter from a lower-order space-variant filter by applying
a global convolution. We note that the space-variant filtering of $f(x,y)$ can generally be written as
\begin{equation*}
\overline{f}(x,y)=\int^{\infty}_{-\infty} \int^{\infty}_{-\infty} f(x-u,y-v) h(x,y; u,v) \ du dv,
\end{equation*}
where $h(x,y;u,v)$ is the space-variant kernel that is indexed by the spatial coordinates $(x,y)$.

\begin{proposition}[Filter decomposition]
\label{split}
Let $h_1(x,y;u,v)$ be a space-variant kernel, and let $h_2(x,y)$ be a convolution kernel. Then the convolution of $f(x,y)$ with $h_2(x,y)$, followed by the space-variant filtering with $h_1(x,y;u,v)$, can be expressed
as a single space-variant filtering:
\begin{equation*}
\overline{f}(x,y)=\int^{\infty}_{-\infty} \int^{\infty}_{-\infty} f(x-u,y-v) h(x,y;u,v) \ du dv,
\end{equation*}
where
\begin{equation*}
h(x,y;u,v)=\int^{\infty}_{-\infty} \int^{\infty}_{-\infty} h_1(x,y; u',v') h_2(u-u',v-v') \ du' dv'.
\end{equation*}
\end{proposition}

Propositions \ref{covariance} and \ref{split} together suggest a simple means of improving the Gaussian approximation for space-variant filtering. 
Suppose that $\mathbf{C}$ is the covariance of the target Gaussian. We split this into two parts, namely,
\begin{equation}
\label{SigmaIso}
\mathbf{C} = \sigma^2 \mathbf{I} + \Delta \mathbf{C}. 
\end{equation}
Note that $\mathbf{C}$ and $\Delta \mathbf{C}$ have the same orientations, that is, $\sigma^2 \mathbf{I}$ possibly alters the elongation, but not the orientation. 
This is confirmed by the fact that $\mathbf{C}$ and $\Delta \mathbf{C}$ have the same set of eigenvectors.  

\begin{algorithm}[!htb]
\caption{Space-variant $O(1)$ Gaussian filtering (better accuracy)}
\label{algorithm1}
\begin{algorithmic}
     \State \textbf{Input}: Image $f(m,n)$, covariance map $\mathbf{C}(m,n)$. 
     \State 1. Set $\sigma^2$ to be half the bound in \eqref{condition2}.
     \State 2. Set $a_1=\cdots=a_4=\sqrt{6} \sigma$.
     \State 3. Pass $a_1,\ldots,a_4$ and $f(m,n)$ to the \texttt{BoxFilter} algorithm, which return $g(m,n)$.
     \State 4. At each pixel, compute $\a(m,n)$ from $\mathbf{C}(m,n) - \sigma^2 \mathbf{I}$.
     \State 5. Pass $\a(m,n)$ and $g(m,n)$ to \texttt{BoxFilter} algorithm, which returns $\overline{f}(x,y)$. 
     \State \textbf{Return}: Filtered image $\overline{f}(x,y)$.
     \end{algorithmic}
\end{algorithm}

The idea is we first convolve the image $f(x,y)$ with an isotropic $\beta(x,y)$ of covariance
$\sigma^2 \mathbf{I}$ (each $a_k$ is $\sqrt{6}\sigma$). This is done using the $O(1)$ algorithm described earlier. We then calculate the residual $\Delta \mathbf{C}$ at each point in the image. This is used to fix $\a$ of
the anisotropic $\beta_{\a}(x,y)$. The scale assignments are then used for the space-variant filtering, again using the $O(1)$ algorithm. The main steps are given in Algorithm \ref{algorithm1}. It is clear that the 
overall algorithm requires $O(1)$ operations per pixel.

\begin{figure}
\centering
\includegraphics[width=1.0\linewidth]{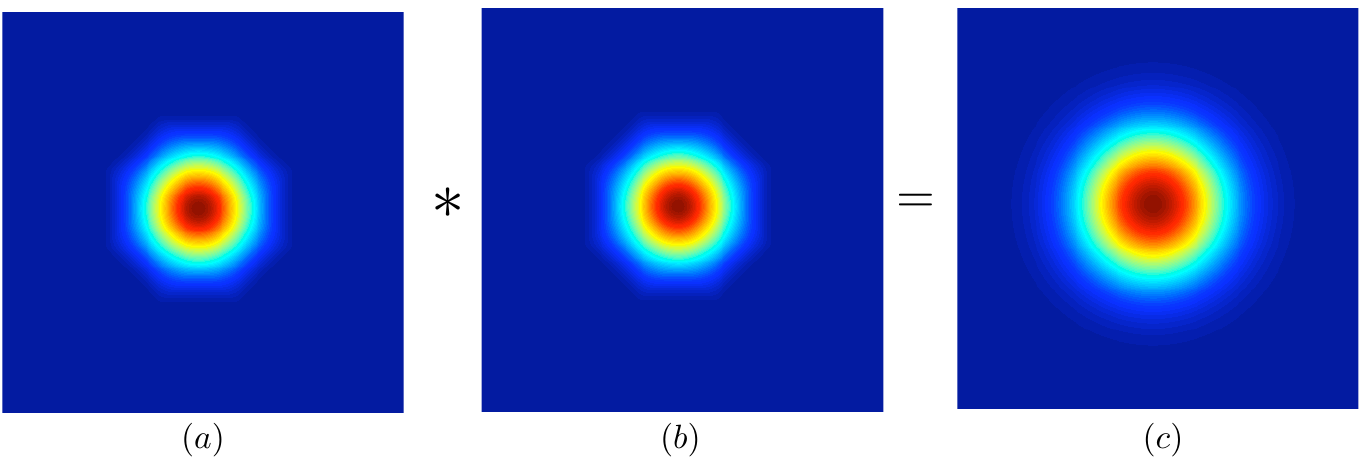}
\includegraphics[width=1.0\linewidth]{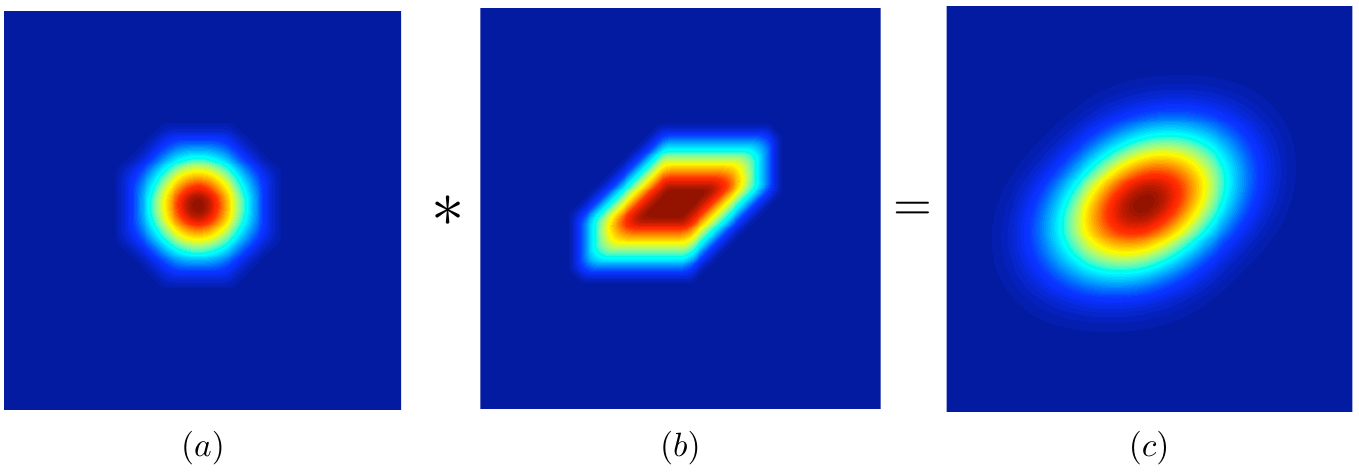}
\caption{This show the steps involved in improving the accuracy of the Gaussian approximation. The isotropic ones are shown on the top panel, while the bottom panel shows the anisotropic ones. In either case,
(a) is the four-directional box spline with covariance $\sigma^2 \mathbf{I}$, and (b) is the four-directional box spline with covariance $\mathbf{C} - \sigma^2 \mathbf{I}$, where $\mathbf{C}$ is the covariance of
the target Gaussian. The resulting box spline approximation is shown in (c), which indeed looks more Gaussian-like than the ones in (b) and (c). See Figure \ref{figure2} for a comparison of the approximations.}
\label{figure1}
\end{figure}

There is a technical point that must be addressed at this point. Namely, $\sigma$ must be so choosen that $\Delta \mathbf{C}$ is again a valid covariance matrix, that is, $\Delta \mathbf{C}$ be positive definite. Now, note that $\Delta \mathbf{C}$ can be written as $\mathbf{Q}^{T}(\Lambda - \sigma^2 \mathbf{I}) \mathbf{Q}$, where $\mathbf{Q}^{T} \Lambda \mathbf{Q}$ is the eigen decomposition of $\mathbf{C}$. 
It follows that a necesaary and sufficient condition for $\Delta \mathbf{C}$ to be a valid covariance matrix is that $\sigma^2$ must be smaller than the minimum eigenvalue of $\mathbf{C}$, that is,
\begin{equation}
 \label{condition1}
 \sigma^2 <  \frac{1}{2} \left( \mathbf{C}_{11}+ \mathbf{C}_{22} -   \sqrt{(\mathbf{C}_{11}-\mathbf{C}_{22})^2+4 \mathbf{C}_{12}^2} \right).
\end{equation}

The other consideration comes from the bound in \eqref{boundE} (this concerns only the anisotropic case), which requires that the   the ratio 
of the eigenvalues of $\Delta \mathbf{C}$ must be with $e(\phi)$, where $\phi$ is the orientation of the target Gaussian. A simple calculation shows that this is equivalent to 
\begin{equation}
\label{condition2}
\sigma^2 <  \frac{1}{2} \left( \mathbf{C}_{11}+ \mathbf{C}_{22} - \left(\frac{e(\phi)+1}{e(\phi)-1}\right)  \sqrt{(\mathbf{C}_{11}-\mathbf{C}_{22})^2+4 \mathbf{C}_{12}^2} \right).
\end{equation}
Since $e(\phi) > 1$, this bound is clearly smaller than the one in \eqref{condition1}, so it suffices to guarantee that \eqref{condition2} holds.

We have empirically observed that, while it is the order of the box splines that dictates the final smoothness, it is the size of $\sigma$ that controls the  
level of approximation. In practice, we have observed that the best approximations are obtained when $\sigma^2$ is roughly $50\%$ of the bound in \eqref{condition2}. 
We will discuss this in detail in the sequel.

\subsection{Improved elongation}

We now address the elongation problem. The elongations achievable by the box spline is minimum exactly mid-way between the axes of the box distributions, namely along 
$22.5^{\circ}, 67.5^{\circ},112.5^{\circ}$, and $157.5^{\circ}$. An intuitive solution to this is to place box distributions along these angles. As mentioned earlier, the problem is that the tangents of these angles are not rational. 
The best we can do in this situation is to approximate the tangents by rational numbers.
In other words, for an given angle $\theta$, the problem reduces to one of finding integers $p$ and $q$ such that $p/q \approx \tan \theta$. Of course, we would also like to keep $p$ and $q$ as small as possible.
Also, for every $\theta$ with $\tan \theta=p/q$, we balance it with the orthogonal angle $\theta+ 90^{\circ}$. This is easily done by setting the tangent of the orthogonal angle to $-q/p$. For example, if we set $p=1$ and $q=2$, 
then we get two orthogonal directions at roughly $26.6^{\circ}$ and $116.6^{\circ}$. The other two
directions are obtained by setting $p=2$ and $q=1$ which gives the orthogonal directions at roughly $63.4^{\circ}$ and $153.4^{\circ}$. Though these are not distributed uniformly over the half-circle, they are off the mark
by only about $4^{\circ}$. To the best of our knowledge, this is the best we can do keeping $p$ and $q$ as small as possible.

This gives us eight directions on the Cartesian grid that are ``almost'' uniformly distributed over the half-circle and whose tangents are rational. The idea then is to place box distributions along all or some of these directions. The higher the number of directions, the better is the 
Gaussian approximation. Unfortunately, as mentioned in the introduction, there are practical difficulties in simultaneously using the  eight directions. Keeping in mind that we already have a solution to improve the accuracy, 
we instead propose the following trick. The idea is to use a pair of four-directional box splines, $\beta_{\a}(x,y)$  with angles $\Theta=\{0^{\circ},45^{\circ},90^{\circ},135^{\circ}\}$,
and $\beta'_{\a}(x,y)$ with angles $\Theta'=\{26.6^{\circ},63.4^{\circ},116.6^{\circ},153.4^{\circ}\}$. We next divide the half-circle into 
two disjoint sectors $S$ and $S'$, where $S$ consists the neighborhood of the former angles, and $S'$ consists of the neighborhood of the latter angles. To approximate a Gaussian with orientation $\phi$, we use 
$\beta_{\a}(x,y)$ if $\phi$ is in $S$, and $\beta'_{\a}(x,y)$ if $\phi$ is in $S'$. This allows us to achieve better elongations than what 
$\beta_{\a}(x,y)$ or $\beta'_{\a}(x,y)$ could achieve by themselves. Of course, we will now require two sets of running sums, one corresponding to each of the box splines. The overall algorithm for space-variant filtering 
is summarized in Algorithm \ref{algorithm2}. 
We note that, as proposed in Section \ref{problemA}, we could smooth the four-directional box splines with a isotropic box spline, if higher accuracy is desired.

\begin{algorithm}{}
\caption{Space-variant $O(1)$ Gaussian filtering (improved elongation)}
\label{algorithm2}
\begin{algorithmic}
     \State 1. \textbf{Input}: Image $f(m,n)$ and covariance map $\mathbf{C}(m,n)$ on $\mathbf{Z}^2$.
     \State 2. Perform running sums on $f(m,n)$ along $\Theta$ to get $g(m,n)$. 
     \State 3. Perform running sums on $f(m,n)$ along $\Theta'$ to get $g'(m,n)$. 
     \State 3. \textbf{At each $(m,n)$ do}
       \State (a) Compute the orientation $\phi$ from $\mathbf{C}(m,n)$.
       \State (b) \textbf{if $\phi \in S$, then} 
	\State    \hspace{5mm} (b1) Compute $\a$ for $\beta_{\a}(x,y)$, and the finite-difference parameters.
	\State    \hspace{5mm} (b2) Apply finite-difference on $g(m,n)$ to get $\overline{f}(m,n)$.  
	\State	  \hspace{4mm} \textbf{else ($\phi \in S'$)} 
	\State    \hspace{5mm} (b1) Compute $\a$ for $\beta'_{\a}(x,y)$, and the finite-difference parameters.
	\State    \hspace{5mm} (b2) Apply finite-difference on $g'(m,n)$ to get $\overline{f}(m,n)$.  
       \State 4. \textbf{Return}: Filtered image $\overline{f}(m,n)$.
\end{algorithmic}
\end{algorithm}

\subsection{Filtering algorithm for $\beta'_{\a}(x,y)$}

We note that the algorithm for space-variant filtering using $\beta'_{\a}(x,y)$ is identical to the one for $\beta_{\a}(x,y)$. For completeness, we describe this in Algorithm \ref{algorithm3}.
Here $x_0,\ldots,x_{15}$ and $y_0,\ldots,y_{15}$ are the coordinates of the finite-difference mesh, and $w_0,\ldots,w_{15}$ are the corresponding weights. The exact values are given in table  1. The shifts $\tau_1$ and $\tau_2$ 
along the  image coordinates are given by
\begin{equation}
\label{shifts}
\tau_1 = \frac{1}{2\sqrt{5}} (2a_1 + a_2 - a_3 - 2a_4) \ \text{and} \ \tau_2 = \frac{1}{2\sqrt{5}}(a_1 + 2a_2 + 2a_3 + a4 - 6\sqrt{5}).
\end{equation}

Note that we have set the scales of the running to $\sqrt{5}$ along each of the four directions. This ensures that no off-grid sample, and hence no interpolation, is required to compute the running sums.
The sample $F(x,y)$ used in the finite-difference is given by
\begin{equation}
\label{interpolation}
F(x,y)=\sum_{|m - x| \leq 3, \ |n - y| \leq 3} \! \! \! g_4(m,n) \beta'_{\mathbf{b}}(x-m,y-n)
\end{equation}
where $\mathbf{b} =(\sqrt{5},\sqrt{5},\sqrt{5},\sqrt{5})$. The box spline $\beta'_{\mathbf{b}}(x,y)$ is supported within the square $[-3,3]^2$, and this is why we have a finite sum in \eqref{interpolation}. 
This can be seen as an interpolation step, where we interpolate the running-sum using the samples of the box spline.

\begin{table*}[!t]
\caption{The positions and weights of the finite-difference mesh used in Algorithm \ref{algorithm3}. Here $\a=(a_1,a_2,a_3,a_4)$ are the scales of the mesh, and $w=1/(a_1a_2a_3 a_4)$ and $a'_i= a_i/\sqrt{5}$.} 
\label{mask_positions} 
\centering
\begin{tabular}{|c|c|c|c|c|c|}
\hline
\bfseries $i$ & \bfseries $(x_i,y_i)$ & \bfseries $w_i$  & $i$ & \bfseries $(x_i,y_i)$ & $w_i$ \\
\hline 
$0$ & $(0,    0)$ &  $+w$ & $8$ & $(-2a'_4, a'_4)$ & $-w$ \\
$1$ & $(2a'_1, a'_1)$ & $-w$ & $9$ & $(2a'_1 - 2a'_4, a'_1 + a'_4)$ & $+w$\\
$2$ & $(a'_2, 2a'_2)$ &  $-w$ & $10$ & $(a'_1 - 2a'_4,  2a'_1 + a'_4)$ & $+w$ \\
$3$ & $(2a_1 + a'_2,  a'_1+2a'_2)$ & $+w$ &  $11$ & $(2a'_1 + a'_2 -2a'_4, a'_1 + 2a'_2 + a'_4)$ & $-w$\\
$4$ & $(-a'_3, 2a'_3)$ & $-w$  & $12$ & $(-a'_3 - 2a'_4, 2a'_3 + a'_4)$ & $+w$ \\
$5$ & $(2a'_1-a'_3, a'_1+2a'_3)$ & $+w$  & $13$ & $(2a'_1 - a'_3 - 2a'_4, a'_1+ 2a'_3 + a'_4)$ & $-w$ \\
$6$ & $(a'_2-a'_3, 2a'_2+2a'_3)$ & $+w$ & $14$ & $(a'_2 - a'_3 - 2a'_4, 2a'_2 + 2a'_3 + a'_4)$ & $-w$ \\
$7$ & $(2a'_1+a'_2-a'_3, a'_1+2a'_2+2a'_3)$ & $-w$ & $15$ & $(2a'_1 + a'_2 - a'_3 - 2a'_4, a'_1 + 2a'_2  + 2a'_3 + a'_4)$ & $+w$ \\
\hline
\end{tabular}
\end{table*}

\subsection{Analytical formulas for $\beta'_{\mathbf{a}}(x,y)$}

The exact formula for the box spline $\beta'_{\mathbf{b}}(x,y)$ in \eqref{interpolation} can be computed analytically. To do so, we note that
this box spline can be expressed as the convolution of two rescaled and rotated box functions. In particular, $\beta'_{\mathbf{b}}(u,v)$ is given by the area of overlap 
between a box function of width $\sqrt{5}$ and height $1/\sqrt{5}$ that is centered at the origin and rotated by $\phi_1$, and a second box function of same width and height that is centered at $(u,v)$ and 
rotated by $\phi_2$,  where $\tan\phi_1 = 1/2$ and $\tan \phi_2 =2$. The overlapping region is a polygon, whose area can be determined by inspection. This can also be found using symbolic computation. 
For example, this can be done in Mathematica by setting
\begin{align*}
\mathtt{f[x,y] \quad := \quad (1/\sqrt{5}) \ UnitBox[(u \cos \phi_1 + v \sin\phi_1)/\sqrt{5}]}  \\
\mathtt{g[x,y] \quad := \quad (1/\sqrt{5}) \ UnitBox[(u \cos \phi_2 + v \sin\phi_2)/\sqrt{5}]}
\end{align*}
and then invoking the command $\mathtt{ Convolve[ \ f[u,v], \ g[u,v], \ \{u,v\}, \ \{x,y\}] }$. The function returned is a piecewise polynomial of second degree that
is compactly supported on a octagonal domain. The equation for each patch is stored in a look-up table.

\begin{algorithm}{}
\caption{Space-variant $O(1)$ filtering using $\beta'_{\a}(x,y)$}
\label{algorithm3}
\begin{algorithmic}
     \State 1. \textbf{Input}: Image $f(m,n)$ and scale map $\a(m,n)$ on $\mathbf{Z}^2$.
     \State 2. Use recursion to compute the following:   \newline
(a) $g_1(m,n) =  \sqrt{5} f(m,n)+g_1(m-1,n-2)$. \newline
(b) $g_2(m,n) = \sqrt{5} g_1(m,n)+g_2(m-2,n-1)$. \newline
(c) $g_3(m,n) = \sqrt{5} g_2(m,n)+g_3(m+1,n-2)$. \newline
(d) $g_4(m,n) = \sqrt{5} g_3(m,n)+g_4(m+2,n-1)$.
     \State 3. \textbf{At each $(m,n)$ do}
       \State (a) Set up $w_i,x_i,y_i (i = 0,1,\ldots,15)$  using table  \ref{mask_positions}, and $\tau_1,\tau_2$ using \eqref{shifts}.
       \State (b) Compute $F_i=F(m+\tau_1-x_i,n+ \tau_2-y_i)$ for $i = 0,1,\ldots,15$ using \eqref{interpolation}.
       \State (c) Set $\overline{f}(m,n)=w_0 F_0+\cdots+w_{15} F_{15}$.
     \State 4. \textbf{Return}: Filtered image $\overline{f}(m,n)$.
\end{algorithmic}
\end{algorithm}

Similar to $\beta_{\a}(x,y)$, we can control the covariance (size, elongation, and orientation) of $\beta'_{\a}(x,y)$ using its scale vector. 
We computed the covariance of $\beta'_{\a}(x,y)$ to be
\begin{equation*}
\frac{1}{60}
\left(\begin{array}{cc} 4a_1^2 + a_2^2 + a_3^2+ 4a_4^2 & 2 (a_1^2+a_2^2-a_3^2-a_4^2) \\
 2(a_1^2+a_2^2-a_3^2-a_4^2)  &  a_1^2 + 4a_2^2 + 4a_3^2 + a_4^2) \end{array}\right).
\end{equation*}
We defined the size, elongation and orientation as in \eqref{geometry}. The final expressions are respectively
\begin{equation*}
\frac{r}{60} , \quad \frac{r + \sqrt{p^2+q^2}}{r -\sqrt{p^2+q^2}}, \ \ \text{and} \ \ \tan^{-1} \left[ \frac{-p+\sqrt{p^2+q^2}}{q}\right],
\end{equation*}
where $p = 3(a_1^2-a_2^2-a_3^2+a_4^2), \ q=4(a_1^2+a_2^2-a_3^2-a_4^2)$, and $r=5(a_1^2+a_2^2+a_3^2+a_4^2)$. 

We note that $a_1,\ldots,a_4$ cannot be uniquely determined from the above three constraints.
As proposed in our earlier paper, we select the box spline that has the minimum kurtosis of all box splines with a given  covariance. 
This turns out to be a one-dimensional optimization problem and can be solved efficiently using a root-finding algorithm. The
kurtosis of $\beta'_{\a}(x,y)$ using the definition in \cite{Kurtosis} turns out to be 
\begin{equation*}
\frac{\kappa}{5}  \left(\begin{array}{cc} 4a_1^4 + a_2^4 + a_3^4+ 4a_4^4 & 2 (a_1^4 + a_2^4 - a_3^4 - a_4^4) \\
 2(a_1^4 + a_2^4 - a_3^4 - a_4^4)  &  a_1^4 + 4a_2^4 + 4a_3^4 + a_4^4) \end{array}\right),
\end{equation*}
where $\kappa$ is the kurtosis of the box function $\mathrm{Box}_1(t)$. To get a number from this matrix, we simply take its Frobenius norm, which is then optimized to get
the scale vector. The steps of the derivation are identical to the one described in \cite{Chaudhury2010}, and therefore we skip the details.

\begin{figure}
\centering
\includegraphics[width=1.0\linewidth]{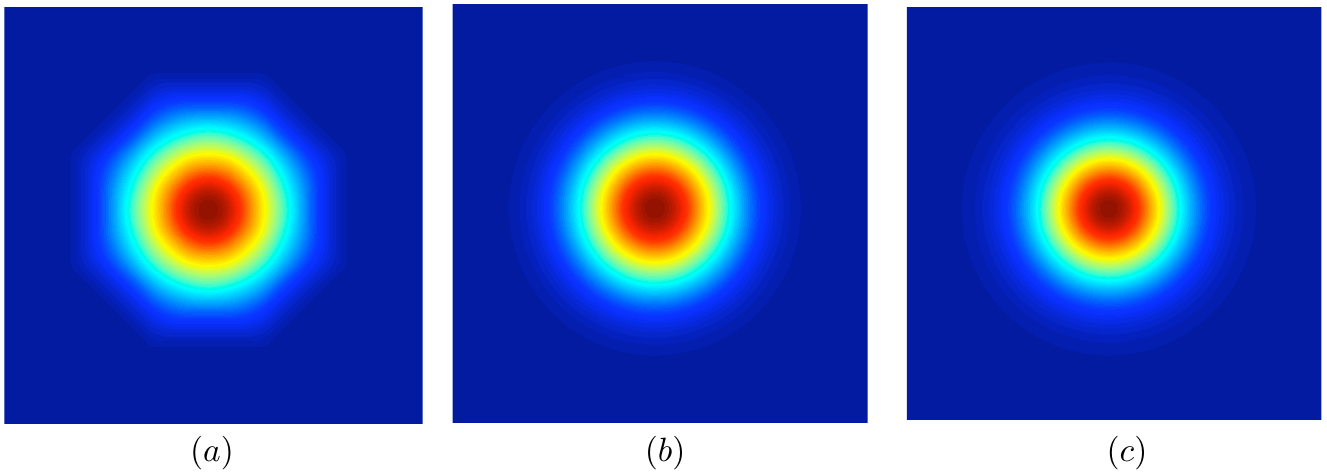}
\includegraphics[width=1.0\linewidth]{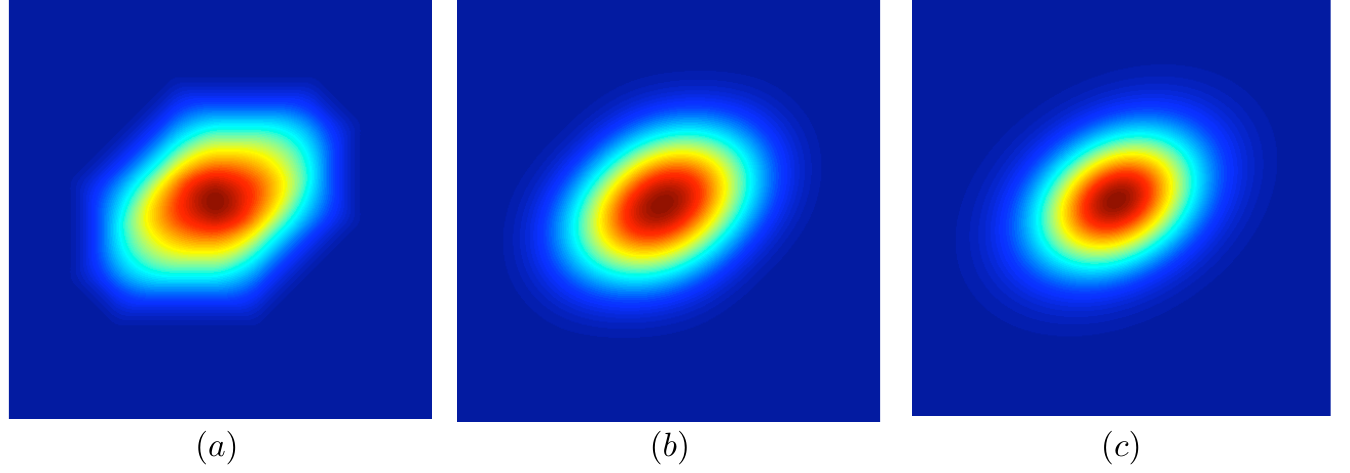}
\caption{In this figure, we compare the target Gaussian (c) with the box spline approximation (a) proposed in \cite{Chaudhury2010} and the one (c) proposed in this work. The isotropic Gaussian has covariance $\mathbf{I}$,
 while the anisotropic one has elongation $3$ and orientation $30^{\circ}$. The maximum pointwise error is within $1\%$ of the peak value for the former and within $2\%$ for the later. A visual comparison confirms the improvement
 in the accuracy -- it is indeed hard to distinguish (b) from (c).}
\label{figure2}
\end{figure}

\section{Experiments}

We now present some results that confirm the improvements suggested in the previous section. First, we try to see the improvement in the accuracy of the Gaussian approximation.
To do so, we consider an isotropic Gaussian with $\mathbf{C}=\mathbf{I}$, and an anisotropic Gaussian with size $1$, elongation $3$, and orientation $30^{\circ}$. In either case, we split
the covariance as $\mathbf{C}=\sigma^2 \mathbf{I} + \Delta \mathbf{C}$. The steps in the
construction are shown in Figure \eqref{figure1}, and the final box splines are compared in Figure \eqref{figure2}. In either case, we have used the four-directional box spline
to approximate the Gaussian with covariance $\sigma^2 \mathbf{I}$, where the four scales were set to $\sqrt{6} \sigma$. To approximate the Gaussian with covariance $\Delta \mathbf{C}$, we used a four-directional 
box spline, but now with unequal scales. The four scales were computed by minimizing the kurtosis of the box spline as explained earlier. 
It is seen from the figures that the final box splines are indeed very good approximations of the target Gaussian. In particular, the maximum pointwise error was within $1\%$ of the peak value for the isotropic one,
while it was within $2\%$ for the anisotropic ones. Note in figure \eqref{figure2} how the convolution helps in ``rounding'' the sharp corners of the four-directional box spline. 

\begin{table}[!htbp]
\caption{Improvement in Gaussian approximation using Algorithm \ref{algorithm1}. The table shows the normalized errors for the \texttt{BoxFilter} in \cite{Chaudhury2010} (second row) 
and the filter proposed in the paper (third row). This is done for certain representative size $(s)$, elongation $(\rho)$, and orientation $(\theta)$ of the target Gaussian.} 
\label{accuracy} 
\centering
\begin{tabular}{|c|ccccccc|}
\hline
$(s,\rho,\theta)$                     & (1,1,0)  &  (5,1,0)   &  (1,4,0)   & (5,4,0)  & (5,3,$\pi/8$)      & (5,8,$\pi/3$)     & (5,5,$\pi/2$)   \\
\hline
Error ($\%$) in \cite{Chaudhury2010}  &  10.8    &   10.8      &   19.1        & 18.7       & 23.9		 & 19.2        & 17.2        \\
\hline
Present error  ($\%$)                 & 4.9      &    4.9     &   15.3        &  14.6      &  20.8             & 15.8	       & 12.6         \\
\hline
Improvement ($\%$)                   &  \bf{54.8}     &    \bf{54.8}     &      \bf{21.6}      &  \bf{18.32} 	    &  \bf{12.7}		 &  \bf{17.7}         &  \bf{26.7}         \\
\hline
\end{tabular}
\end{table}

To get a better understanding of the  improvement in the accuracy, we compare the approximation error between the target Gaussian and the two box splines. We do this for various sizes, elongations, and orientations 
of the target Gaussian $g(x,y)$. We use the ratio $\lVert f - g \rVert_2/\lVert g \rVert_2$ as the normalized error 
between the approximating box spline $f(x,y)$ and $g(x,y)$, where $\lVert f \rVert_2$ is the $L_2$ norm of a function. We computed the error from the analytical formulas of the Gaussian 
and the box splines using numerical integration. The results are reported in table  \ref{accuracy}. In the first two instances (isotropic setting), note that there is almost a $40\%$ improvement in the approximation
(this is the case for any size). The improvement is less dramatic, but nonetheless appreciable, for the remaining five instances (anisotropic setting). We found that this is between $10\% - 30\%$ for the anisotropic box 
splines in general. The improvement is on the higher side when the orientation is close to one of the four axes of the box spline.

We now discuss the choice of $\sigma$ in \eqref{SigmaIso} used in the above experiments. As noted earlier, we have observed that the best results are obtained when $\sigma^2$ for the isotropic box spline is roughly 
$50\%$ of the bound in \eqref{condition2}. The variation in the approximation accuracy is shown in table  \ref{sigma} for a particular Gaussian. Exhaustive experiments across various settings have shown that 
this happens consistently across different size, elongation, and orientation. The explanation for this is as that very little smoothing takes place when $\sigma$ is small, 
which explains the marginal improvement in accuracy. On the other hand, large values of $\sigma$ result in over-smoothing, where the isotropic box spline dominates the overall anisotropy -- the contribution from 
the anisotropic part is less significant. Following these observations, we set $\sigma^2$ to be $50\%$ of the bound in \eqref{condition2}.

\begin{table}[!htbp]
\caption{Effect of $\sigma$ in \eqref{SigmaIso} on the Gaussian approximation. The table shows the difference between (a) the approximation error between a fixed Gaussian
($s = 5, \rho = 3, \theta = \pi/4$) and the \texttt{BoxFilter}, and (b) the error between the same Gaussian and the present filter. The best results are obtained when $\sigma^2$
is roughly $50\%$ of the bound in \eqref{condition2}.} 
\label{sigma} 
\centering
\begin{tabular}{|c|cccccccc|}
\hline
$\%$ of the bound in \eqref{condition2}       & 10    & 20     &  30   &  40     &  \bf{50}   &  60    & 70    & 80  \\
\hline
Improvement ($\%$)                            & 3.5   & 12.1   & 22.3  &  29.9   & \bf{33.3}  & 33.13   & 30.9  &  28.8  \\
\hline
\end{tabular}
\end{table}

We now study the improvement in elongation obtained using Algorithm \ref{algorithm2}. While there is an explicit formula for the bound of $\beta_{\a}(x,y)$, we computed the bound of $\beta'_{\a}(x,y)$ empirically. 
Table \ref{boundElongation} gives a comparison of the maximum elongations achievable using $\beta_{\a}(x,y)$ and $\beta'_{\a}(x,y)$ as against 
that achievable using only $\beta_{\a}(x,y)$. 
In the table, we give the bounds on the elongation for orientations in the sector $[0^{\circ}, 45^{\circ}]$, spanned between two neighboring axes of $\beta_{\a}(x,y)$. While large elongations are achievable using  $\beta_{\a}(x,y)$ in the
the caps $[0^{\circ}, 5^{\circ}]$ and $[40, 45^{\circ}]$, the bound falls off quickly outside these caps. In particular, the bound falls to a minimum of $5.8$ at $22.5^{\circ}$.
The second box spline $\beta'_{\a}(x,y)$, with its axis along $26.6^{\circ}$, helps us improve the elongation over $[5^{\circ},40^{\circ}]$. For example, the bound at $22.5^{\circ}$ has now increased to $10.8$, and remains fairly
high in the cap $[20^{\circ}, 40^{\circ}]$.
However, a minimum of $8.2$ is reached at $13.3^{\circ}$, exactly midway between the axes of $\beta_{\a}(x,y)$ and $\beta'_{\a}(x,y)$. This is nevertheless better than the previous low of $5.8$.
We have evidence that the use of a third box spline with axes close to $13.3^{\circ}$ can improve the minimum bound to $15.2$. We note that the use of multiple box splines increases the run time of the 
\texttt{BoxFilter} by only $10\%-15\%$. The added computations are required only from the running sums, which is the fastest part of the algorithm.
The finite-difference part of the algorithm remains the same. For every orientation, one simply has to choose the correct running sum and the associated finite-difference.

\begin{table}[!htbp]
\caption{Comparison of the elongations achievable in \cite{Chaudhury2010} and using Algorithm \ref{algorithm2}. The table shows the bounds on the elongation at different orientations using a single box spline filter 
(second row) and a pair of box spline (third row). The new bound is the maximum of the bounds of the two box splines.} 
\label{boundElongation} 
\centering
\begin{tabular}{|c|cccccccccc|}
\hline
Orientation      & $0^{\circ}$       & $5^{\circ}$          & $13.3^{\circ} $    & $ 20^{\circ}$  & $ 22.5^{\circ}$ &  $25^{\circ}$  & $ 26.6^{\circ}$  &  $30^{\circ}$   &  $40^{\circ}$    & $45^{\circ}$ \\
\hline 
Previous bound   & $\bf{\infty}$     & 13.6                &  6.8               &  5.9            & \bf{5.8}       & 5.9               &  6.0             &  6.1            & 13.6             &  $\bf{\infty}$  \\
\hline
New bound       &  $\bf{\infty}$     & 13.6                & \bf{8.2}           & 9.5             & 10.8           & 30.1              &  $\bf{\infty}$   &  28.8           & 13.6             &  $\bf{\infty}$ \\
\hline
\end{tabular}
\end{table}

\begin{figure}
\centering
\includegraphics[width=1.0\linewidth]{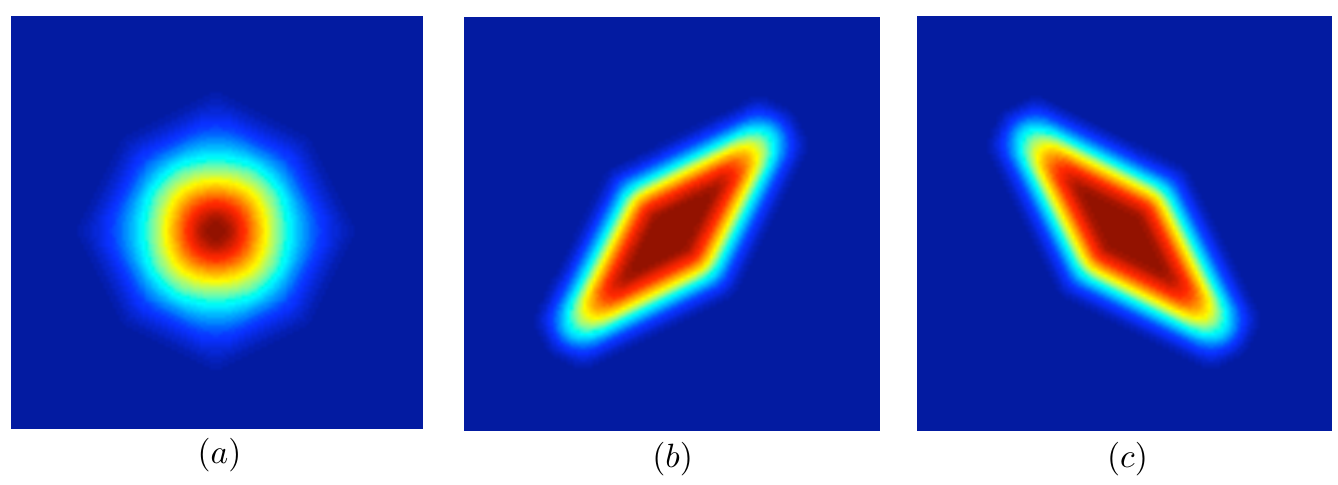}
\caption{This shows the Gaussian-like blobs obtained by applying Algorithm \ref{algorithm3} to an impulse image. (a) is the isotropic response obtained using equal scales. (b) and (c) are the anisotropic responses obtained 
using unequal scales. The latter are oriented along $45^{\circ}$ and $135^{\circ}$, and have elongation $3$.}
\label{figure6}
\end{figure}

We have implemented a basic version of Algorithm \ref{algorithm1} as an ImageJ plugin. The Java code can be found at www.math.princeton.edu/$\sim$kchaudhu/code.html. Some of the results in this paper can be reproduced 
using this plugin. 
The typical run time on a $256 \times 256$ image (Intel quad-core 2.83 GHz processor) was around $300$ milliseconds.  As expected, the run time was approximately the same for filters of different shape and size. 
On the other hand, we have tested the formulas for the box spline $\beta_{\a}(x,y)$ by implementing Algorithms \ref{algorithm2} in Matlab. The typical run time was around $600$ milliseconds 
on a $256 \times 256$ image. 
The results obtained on a impulse image are shown in Figure \ref{figure6}.
In the future, we plan to improve the run time by implementing the algorithms in Java.

\section{Discussion}

In this work, we showed that the performance of the box spline filters in \cite{Chaudhury2010} can be improved, while preserving the $O(1)$ complexity
of the orignal algorithm. 
In particular, we showed how the accuracy of the Gaussian approximation can be improved by pre-filtering the image with an isotropic Gaussian of appropriate size. This idea is in fact quite generic,
and can be used to improve the performance of any algorithm for space-variant filtering.
We also showed that better elongations can be achieved using a pair of four-directional box splines instead of one. The present work suggests that both the accuracy and the elongation can
be dramatically improved, at a small extra cost, by using a bank of four-directional box splines (ideally, three or four) whose axes are evenly distributed over the half-circle. We will investigate this
possibility in the future.

\section{Acknowledgment}

The authors would like to thank Professor Michael Unser and Dr. Arrate Munoz-Barrutia for valuable discussions. This work was supported by the Swiss National Science Foundation under grant PBELP2-$135867$.


\begin{thebibliography}{9}

\bibitem{Viola2001} P. Viola and M. Jones, ``Rapid object detection using a boosted cascade of simple features,'' IEEE Conference on Computer Vision and Pattern Recognition, pp. 511-518, 2001.

\bibitem{Crow1984} F.C. Crow, ``Summed-area tables for texture mapping,'' ACM Siggraph, vol. 18, pp. 207-212, 1984.


\bibitem{Chaudhury2010} K.N. Chaudhury, A. Munoz-Barrutia, M. Unser, ``Fast space-variant elliptical filtering using box splines,'' IEEE Transactions on Image Processing, vol. 19, no. 9, pp. 2290-2306, 2010.

\bibitem{Heckbert1986} P. S. Heckbert, ``Filtering by repeated integration,'' International Conference on Computer Graphics and Interactive Techniques, vol. 20, no. 4, pp. 315-321, 1986.

\bibitem{ChaudhuryArxiv} K.N. Chaudhury, ``A radial version of the Central Limit Theorem,'' arXiv:1109.2227v1 [cs.IT].

\bibitem{Wetzstein2011} G. Wetzstein, D. Lanman, W. Heidrich, and R. Raskar, ``Layered 3D: tomographic image synthesis for attenuation-based light field and high dynamic range displays,'', ACM Transactions on Graphics, vol. 30, no. 4, pp. 1-12, 2011.

\bibitem{deBoor} C. de Boor, K. Hollig, and S. Riemenschneider, Box Splines, Springer-Verlag, 1993.


\bibitem{Chaudhury2011} K.N. Chaudhury, ``Constant-time filtering using shiftable kernels,'' IEEE Signal Processing Letters, 2011.



\bibitem{Weickert1994} J. Weickert, ``Theoretical foundations of anisotropic diffusion in image processing,'' Theoretical foundations of computer vision, vol. 11, pp. 221-236, 1994.

\bibitem{Arrate2002} A. Munoz-Barrutia, R. Ertle, and M. Unser, ``Continuous wavelet transform with arbitrary scales and $O(N)$ complexity,'' Signal Processing, vol. 82, no. 5, pp. 749-757, 2002.


\bibitem{ChaudhuryISBI2010} K.N. Chaudhury, Zs. Puspoki, A. Munoz-Barrutia, D. Sage, and M. Unser, ``Fast detection of cells using a continuously scalable {M}exican-hat-like template,'' IEEE International Symposium on Biomedical Imaging, pp. 1277-1280, 2010.



\bibitem{Arrate2010} A. Munoz-Barrutia, X. Artaechevarria, and C. Ortiz-de Solorzano, ``Spatially variant convolution with scaled B-splines,'' IEEE Transactions on Image Processing, vol. 19, no. 1, pp. 11-24, 2010.


\bibitem{Kurtosis} A. Tkacenko and P.P. Vaidyanathan, ``Generalized kurtosis and applications in blind equalization of MIMO channels,'' Conference Record of the Thirty-Fifth Asilomar Conference on Signals, Systems and Computers 1 
, 742 - 746, 2001.

\end{thebibliography}
\end{document}